\documentclass{article}

\usepackage{arxiv}

\usepackage[linesnumbered]{algorithm2e}
\usepackage{algpseudocode}
\usepackage{amsfonts}
\usepackage{amsmath}
\usepackage{booktabs}
\usepackage{color}
\usepackage{graphicx}
\usepackage{makecell}
\usepackage{tabularx}
\usepackage{subcaption}
\usepackage{amssymb}
\usepackage{url}

\newcommand{\figccwidth}[4]{
    \begin{figure*}[tb]\centering
    \includegraphics[width=#4]{./fig/#1}
    \caption{#2}
    \label{#3}\end{figure*}
    {}}

\usepackage{amsmath,amsfonts,bm}

\def\eqref#1{equation~\ref{#1}}

\def\1{\bm{1}}

\DeclareMathAlphabet{\mathsfit}{\encodingdefault}{\sfdefault}{m}{sl}
\SetMathAlphabet{\mathsfit}{bold}{\encodingdefault}{\sfdefault}{bx}{n}

\title{Dynamic DropConnect: Enhancing Neural Network Robustness through Adaptive Edge Dropping Strategies}

\author{
 Yuan-Chih Yang, Hung-Hsuan Chen\thanks{Corresponding author
 } \\
 National Central University \\
 \texttt{ericabd888@gmail.com, hhchen1105@acm.org}
 }

\date{}

\begin{document}

\maketitle

\begin{abstract}

Dropout and DropConnect are well-known techniques that apply a consistent drop rate to randomly deactivate neurons or edges in a neural network layer during training. This paper introduces a novel methodology that assigns dynamic drop rates to each edge within a layer, uniquely tailoring the dropping process without incorporating additional learning parameters. We perform experiments on synthetic and openly available datasets to validate the effectiveness of our approach. The results demonstrate that our method outperforms Dropout, DropConnect, and Standout, a classic mechanism known for its adaptive dropout capabilities. Furthermore, our approach improves the robustness and generalization of neural network training without increasing computational complexity. The complete implementation of our methodology is publicly accessible for research and replication purposes at \url{https://github.com/ericabd888/Adjusting-the-drop-probability-in-DropConnect-based-on-the-magnitude-of-the-gradient/}.

\end{abstract}

\section{Introduction}

Dropout~\cite{hinton2012improving,srivastava2014dropout} and DropConnect~\cite{wan2013regularization} are prominent techniques designed to mitigate overfitting in deep neural networks. Dropout functions by independently zeroing each neuron within a layer with a predetermined fixed probability $p$. In contrast, DropConnect takes a slightly different approach by randomly eliminating an edge within the network with a fixed probability $p$. This makes DropConnect a generalization of Dropout; specifically, removing a single neuron as performed in Dropout equates to eliminating all incoming and outgoing edges associated with that neuron, which DropConnect facilitates.

Both Dropout and DropConnect uniformly apply a fixed dropping rate across all neurons or edges within a layer. However, this universal dropping rate might not always represent the optimal strategy; ideally, a model should utilize available data-driven insights to tailor the dropping rate for each individual edge or neuron based on their specific characteristics.

To address this, we introduce a novel methodology termed DynamicDropConnect (DDC). This approach dynamically assigns a drop probability to each edge based on the magnitude of the gradient associated with that edge. The underlying principle is that edges with larger gradients are crucial for learning and should be retained, whereas it might be acceptable to omit those with minimal impact on the model's output occasionally.

\figccwidth{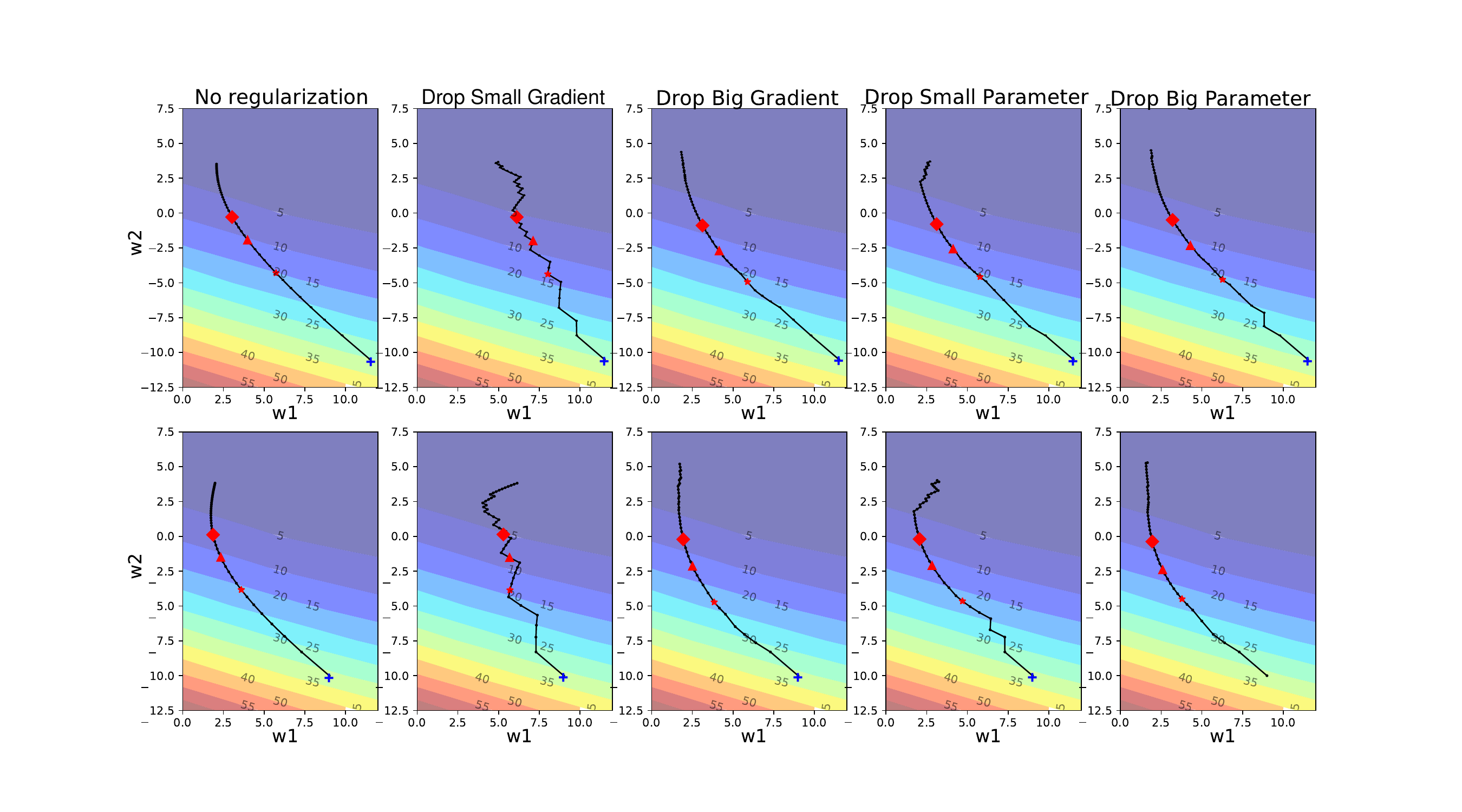}{The contour plot of the loss and the parameter update process. The two rows represent two sets of initial values and their updating process. The blue pluses denote the initial values. The red stars, triangles, and diamonds represent the values of $w_1$ and $w_2$ after training for 7, 13, and 19 epochs. The method that tends to drop the edges with small gradients (the second column) reaches a small error faster.}{fig:linreg-contours}{1.\textwidth}

DDC offers several advantages over other methods such as Standout~\cite{ba2013adaptive}, which also employs a dynamic dropping rate. Firstly, DDC does not require any additional learning parameters, which simplifies the model architecture and reduces memory requirements during training. Secondly, DDC provides a more deterministic and transparent approach to deciding the dropping rate, as opposed to the potentially opaque and unpredictable learning processes used by methods like Standout. Moreover, empirical evidence from our experiments demonstrates that DDC achieves superior accuracy compared to Standout.

Our experimental analysis extends beyond theoretical discussions. We conduct rigorous experiments on both synthetic and real open datasets to validate the efficacy of our methodology. Figure~\ref{fig:linreg-contours} illustrates the trajectory of parameter values and their corresponding losses using a synthetic dataset. The first column in the figure depicts a model without any regularization (labelled as ``No regularization'' in Figure~\ref{fig:linreg-contours}, whereas the second column shows the impact of training the model by preferentially dropping edges associated with smaller gradients (``Drop Small Gradient''). For comparative purposes, we also trained the same model using different variants (``Drop Big Gradient'', ``Drop Small Parameter'', and ``Drop Big Parameter''). The plots clearly demonstrate that training a linear regressor using our methodology, which preferentially drops edges with smaller gradients, reaches minimal losses more quickly than other compared methods. This encouraging result has motivated us to further explore the performance of our method when applied to more complex networks on open datasets.

The rest of the paper is organized as follows: Section~\ref{sec:rel-work} reviews related studies. Section~\ref{sec:method} details the proposed method. Section~\ref{sec:exp} evaluates various dropping strategies using different network architectures on diverse datasets. Finally, Section~\ref{sec:conc} concludes our work and outlines potential future research directions.
\section{Related Work} \label{sec:rel-work}

Overfitting occurs when a model excels during the training phase but performs inadequately on unseen test data. A typical approach to counter overfitting involves incorporating regularization terms into the objective function, such as the L1 or L2 norms of the learnable parameters. These regularization terms are designed to encourage the development of simpler relationships between features and targets, thus mitigating the risk of overfitting. Early stopping is another frequently adopted strategy to prevent overfitting. This technique involves ceasing the parameter updates once the loss on the validation set begins to show signs of increase, thereby avoiding the overfitting of the training data. Additional well-established methods to combat overfitting include data augmentation~\cite{krizhevsky2017imagenet}, employing model ensembles~\cite{ke2017lightgbm}, and modifying the architecture of neural networks by reducing their depth or width. These methods have been proven to be effective in enhancing the generalizability of models across unseen datasets.

Dropout is another widely recognized method specifically utilized in neural networks to curb overfitting~\cite{srivastava2014dropout}. During training, Dropout randomly deactivates a subset of neurons, ensuring that the neural network does not become overly dependent on any particular set of neurons. This technique effectively simulates the use of multiple neural network models and aggregates their predictions, which enhances the robustness of the model. Notably, the random neuron deactivation is only applied during the training phase and not during inference. To maintain consistent expected values between training and inference, Inverted Dropout is commonly utilized.

DropConnect extends the idea of Dropout by randomly omitting connections, or edges, between neurons during the training phase. This approach is similar to Dropout in that masking a neuron in Dropout is analogous to masking both the incoming and outgoing connections of that neuron in DropConnect. Empirically, both Dropout and DropConnect often result in models with comparable performance levels.

From a Bayesian perspective, Dropout can also be interpreted as performing an approximated Bayesian inference in deep Gaussian processes~\cite{gal2016dropout}. Notable techniques within this framework include Monte Carlo Dropout~\cite{gal2016dropout}, fast dropout~\cite{wang2013fast}, and variational dropout~\cite{kingma2015variational}, each offering unique approaches to implementing Dropout within a Bayesian inference context.

Some recent studies have explored the idea of varying the dropping rate throughout the training process. A notable example is Standout~\cite{ba2013adaptive}, which dynamically adjusts the dropping rate based on the values of the model parameters. Unlike other variations of Standout, which often necessitate additional learning parameters, our approach simplifies the process by avoiding extra learning mechanisms. This not only reduces training time and memory requirements but also provides a more transparent method for determining dynamic dropping rates. Furthermore, our research has explored the efficacy of utilizing gradient values rather than parameter values for adjusting drop rates, which has shown promising results in our experimental evaluations.

For a comprehensive review of Dropout and its various adaptations, please refer to the survey by Labach et al.~\cite{labach2019survey}.
\section{Methodology} \label{sec:method}

This section introduces DropConnect as the background knowledge, followed by our methodology to assign dynamic dropping rates.

\subsection{Preliminary: DropConnect}

Let $l \in \left\{1, \ldots ,L\right\}$ be the $L$ hidden layers in a neural network and $\boldsymbol{y}^{(l)}$ be the output of layer $l$ (and therefore the input of layer $l+1$), each layer of a neural network transforms $\boldsymbol{y}^{(l-1)}$ to $\boldsymbol{y}^{(l)}$ by Equation~\ref{eq:normal-forward}.

\begin{equation} \label{eq:normal-forward}
\boldsymbol{y}^{(l)} = f_l\left(\boldsymbol{z}^{(l)}\right) = f_l\left(\boldsymbol{W}^{(l)} \boldsymbol{y}^{(l-1)}\right),
\end{equation}
where $\boldsymbol{W}^{(l)}$ is the set of parameters in layer $l$, and $f_l(\cdot)$ is the activation function of the same layer.

DropConnect assigns a universal dropping rate $p$ such that each edge has a probability $p$ of being turned off during training. Therefore, a neural network, with DropConnect, transforms $\boldsymbol{y}^{(l-1)}$ to $\boldsymbol{y}^{(l)}$ by the equation below during training.

\begin{equation} \label{eq:dropconnect-forward}
\boldsymbol{y}^{(l)} = f_l\left(\boldsymbol{z}^{(l)}\right) =
f_l\left( \left( (1-\boldsymbol{M}^{(l)}) \odot \boldsymbol{W}^{(l)}\right) \boldsymbol{y}^{(l-1)}\right),
\end{equation}
where $\boldsymbol{M}^{(l)} = \left[m_{i,j}\right]$ is a $(0,1)$-matrix whose shape is the same as $\boldsymbol{W}^{(l)}$; each entry $m_{i,j}$ is sampled from a Bernoulli distribution with a fixed hyper-parameter $p$, and $\odot$ performs the Hadamard product (i.e., element-wise product) of matrices.  Thus, $\boldsymbol{M}^{(l)}$ is a mask matrix that decides which parameters in $\boldsymbol{W}^{(l)}$ (edges) to omit. 

\subsection{DDC -- Mask Generation} \label{sec:gen-mask}

\RestyleAlgo{ruled}
\begin{algorithm}[tb]
\caption{Generating a mask for $\boldsymbol{W}^{(l)}$}\label{alg:ddc-gen-mask}
\SetKwInOut{Input}{Input}
\SetKwInOut{Output}{Output}
\Input{
gradients $\boldsymbol{G}^{(l)} = \left[g_{i,j}^{(l)}\right] \in \mathcal{R}^{n_1 \times n_2}$,
\\ hyperparameters $p$, $p_g$, and $\tau$}
\Output{mask $\widetilde{\boldsymbol{M}}^{(l)} = \left[\widetilde{m}_{i,j}^{(l)}\right] \in \mathcal{R}^{n_1 \times n_2}$}
\For{$i \gets 1$ \KwTo $n_1$}{
    \For{$j \gets 1$ \KwTo $n_2$}{
        $v_{i,j}^{(l)} \gets \left\lvert g_{i,j}^{(l)}\right\rvert$\;
    }
}
$\mu^{(l)} \gets \dfrac{\sum_{i=1}^{n_1} \sum_{j=1}^{n_2} v_{i,j}^{(l)}}{n_1 \times n_2}$\;
$\sigma^{(l)} \gets \dfrac{\sum_{i=1}^{n_1} \sum_{j=1}^{n_2} (v_{i,j}^{(l)} - \mu^{(l)})^2}{n_1 \times n_2}$\;
\For{$i \gets 1$ \KwTo $n_1$}{
    \For{$j \gets 1$ \KwTo $n_2$}{
        $z_{i,j}^{(l)} \gets \dfrac{v_{i,j}^{(l)} - \mu^{(l)}}{\sigma^{(l)}}$\;
        Update $q_{i,j}^{(l)}$ by Equation~\ref{eq:candidate-drop-prob}\;
        Update $p_{i,j}^{(l)}$ by Equation~\ref{eq:drop-rate}\;
        $\widetilde{m}_{i,j}^{(l)} = \text{Bernoulli}(p_{i,j}^{(l)})$\;
    }
}
\end{algorithm}

The proposed DDC advances DropConnect by allowing variable dropping probabilities for distinct neural network edges. 

We create a mask matrix, $\widetilde{\boldsymbol{M}}^{(l)}$ to drop edges. Each entry $\widetilde{m}_{i,j}^{(l)}$ in $\widetilde{\boldsymbol{M}}^{(l)}$ is sampled from a Bernoulli distribution with a unique parameter, $p_{i,j}^{(l)}$. An edge is omitted if $\widetilde{m}_{i,j}^{(l)} = 1$.

Algorithm~\ref{alg:ddc-gen-mask} shows a pseudocode to generate $\widetilde{\boldsymbol{M}}^{(l)}$ during training. For every edge $e_{i,j}^{(l)}$ that bridges neuron $i$ at layer $l-1$ and neuron $j$ at layer $l$, the algorithm decides the value of the corresponding mask cell $\widetilde{m}_{i,j}^{(l)}$ based on its current gradient $g_{i,j}^{(l)}$ at every training iteration. The algorithm first computes the absolute gradient value of $g_{i,j}^{(l)}$, represented as $v_{i,j}^{(l)}$ (line 1 to line 5). DDC needs gradient normalization to harmonize the gradient magnitudes across distinct layers. Therefore, the mean and standard deviation of the set of $v_{i,j}^{(l)}$ values for layer $l$ are calculated (line 6 and line 7). Subsequently, a z-score normalization is performed on $v_{i,j}^{(l)}$, yielding normalized gradient magnitudes $z_{i,j}^{(l)}$-s.
DDC then calculates $q_{i,j}^{(l)}$ for each edge based on the normalized gradient magnitude $z_{i,j}^{(l)}$.  This candidate dropping probability positively correlates with the edge's final dropping probability. Eventually, we define the candidate dropping probability based on Equation~\ref{eq:candidate-drop-prob}, so the model tends to keep the edge with a larger gradient magnitude.

\begin{equation} \label{eq:candidate-drop-prob}
q_{i,j}^{(l)} \gets \begin{cases}
            1 - \sigma\left(z_{i,j}^{(l)}\right) & \text{if } 1 - \sigma\left(z_{i,j}^{(l)}\right) \geq \tau \\
            0 & \text{otherwise},
        \end{cases}
\end{equation}
where $\sigma()$ is the sigmoid function.  Since $\sigma()$ is monotonically increasing and the output is always between 0 and 1, a larger $z_{i,j}^{(l)}$ results in a smaller candidate dropping probability. We set hyperparameter $\tau$ to $0.5$. So, half of the $q_{i,j}^{(l)}$-s are zero on average because the expected value of $\sigma\left(z_{i,j}^{(l)}\right)$ is 0.5.

The final dropping rate $p_{i,j}^{(l)}$ for each edge $e_{i,j}^{(l)}$ is a function of three variables: the candidate dropping probability $q_{i,j}^{(l)}$, a base dropping probability $p$ (a hyper-parameter), and a gradient unit dropping rate $p_g$ (another hyper-parameter). We limit the value of $p_{i,j}^{(l)}$ to be no greater than 1 to ensure $p_{i,j}^{(l)}$ as a proper probability value. The computation of $p_{i,j}^{(l)}$ is shown in Equation~\ref{eq:drop-rate}.

\begin{equation} \label{eq:drop-rate}
p_{i,j}^{(l)} \gets \begin{cases}
            p + p_g \times q_{i,j}^{(l)} & \text{if } p + p_g \times q_{i,j}^{(l)} \le 1 \\
            1 & \text{otherwise}.
        \end{cases}
\end{equation}

Finally, the DDC generates each $\widetilde{m}_{i,j}^{(l)}$ in $\widetilde{\boldsymbol{M}}^{(l)}$ by sampling from a Bernoulli distribution with the parameter $p_{i,j}^{(l)}$ (line 13).

\subsection{DDC -- Training Re-calibration and Testing} \label{sec:ddc-forward}

\begin{algorithm}[tb]
\caption{One-layer forward re-calibration during training}\label{alg:one-layer-forward}
\SetKwInOut{Input}{Input}
\SetKwInOut{Output}{Output}
\Input{weights $\boldsymbol{W}^{(l)} = \left[w_{i,j}^{(l)}\right] \in \mathcal{R}^{n_1 \times n_2}$,\\
gradients $\boldsymbol{G}^{(l)} = \left[g_{i,j}^{(l)}\right] \in \mathcal{R}^{n_1 \times n_2}$,\\ 
outputs from previous layer $\boldsymbol{y}^{(l-1)} \in \mathcal{R}^{n_2 \times 1}$}
\Output{$\boldsymbol{y}^{(l)} \in \mathcal{R}^{n_1 \times 1}$}
Compute $\widetilde{\boldsymbol{M}}^{(l)}$ by Algorithm~\ref{alg:ddc-gen-mask} using $\boldsymbol{W}^{(l)}$ and $\boldsymbol{G}^{(l)}$\;
$\widetilde{\boldsymbol{W}}^{(l)} \gets \left(1-\widetilde{\boldsymbol{M}}^{(l)} \right) \odot \boldsymbol{W}^{(l)}$\;
Compute $r^{(l)}$ by Equation~\ref{eq:para-dropping-rate}\;
Compute $\boldsymbol{y}^{(l)}$ by Equation~\ref{eq:compensated-output}\;
\end{algorithm}

Since $\widetilde{\boldsymbol{M}}^{(l)}$ is the mask for the parameters in layer $l$, the effective parameters in this layer go from $\boldsymbol{W}^{(l)}$ to $(1-\widetilde{\boldsymbol{M}}^{(l)}) \odot \boldsymbol{W}^{(l)}$. The dropping mechanism is only applied in training but not in prediction. However, if we set the value of $\boldsymbol{y}^{(l)}$ as $f_l\left((1-\widetilde{\boldsymbol{M}}^{(l)}) \odot \boldsymbol{W}^{(l)} \boldsymbol{y}^{(l-1)}\right)$ in training but predict $\boldsymbol{y}^{(l)}$ as $f_l\left(\boldsymbol{W}^{(l)} \boldsymbol{y}^{(l-1)}\right)$ in inference, the learned $\boldsymbol{W}^{(l)}$ from training cannot be used directly in the tests.  To fix the inconsistency, we need to recalibrate $(1-\widetilde{\boldsymbol{M}}^{(l)}) \odot \boldsymbol{W}^{(l)}$ during training, so that the learned $\boldsymbol{W}^{(l)}$ in training can be used directly during inference by $f_l\left(\boldsymbol{W}^{(l)} \boldsymbol{y}^{(l-1)}\right)$.

Since $\widetilde{\boldsymbol{M}}^{(l)}$ is a $(0,1)$-matrix, $\sum_{i=1}^{n_1} \sum_{j=1}^{n_2} \widetilde{m}_{i,j}^{(l)}$ is the number of 1-s in $\widetilde{\boldsymbol{M}}^{(l)}$.  So, we can compute the expected value of the dropping rate of layer $l$ by Equation~\ref{eq:para-dropping-rate}.

\begin{equation}\label{eq:para-dropping-rate}
r^{(l)} = \frac{\sum_{i=1}^{n_1}\sum_{j=1}^{n_2} \widetilde{m}_{i,j}^{(l)}}{n_1 \times n_2},
\end{equation}
where $n_1$ and $n_2$ are the number of rows and the number of columns of the masking matrix $\widetilde{\boldsymbol{M}}^{(l)}$.  

We re-calibrate the output of layer $l$ during training such that the learned weights $\boldsymbol{W}$ can be used during inference. The re-calibration is achieved by dividing the output by the keep rate:

\begin{equation} \label{eq:compensated-output}
\boldsymbol{y}^{(l)} \gets f_l\left(\widetilde{\boldsymbol{W}}^{(l)} \boldsymbol{y}^{(l-1)} \times \dfrac{1}{1 - r^{(l)}}\right).
\end{equation}

The mask matrix $\widetilde{\boldsymbol{M}}^{(l)}$ and the masked parameter matrix $\widetilde{\boldsymbol{W}}^{(l)}$ are only used in training.  At inference, we use only the unmasked parameter matrix $\boldsymbol{W}^{(l)}$ for forwarding: $\boldsymbol{y}^{(l)} \gets f_l\left(\boldsymbol{W}^{(l)} \boldsymbol{y}^{(l-1)}\right)$.

Algorithm ~\ref{alg:one-layer-forward} shows one-layer forward re-calibration in training.

\section{Experiments} \label{sec:exp}

This section introduces the experiment settings and results. All the models are implemented by PyTorch and trained on the NVIDIA GTX 3090. We conducted experiments based on one synthetic dataset and four open datasets: MNIST, CIFAR-10, CIFAR-100, and NORB.  
We split the labeled instances for each open dataset into the training set, the validation set, and the test set.  
Detailed settings, such as parameter initialization, learning rate, and batch size, are included in the experimental code.

\subsection{Experiments on the Synthetic Dataset} \label{sec:syn-data}

\begin{figure}[tb]
     \centering
     \begin{subfigure}{0.48\textwidth}
         \centering         \includegraphics[width=\textwidth]{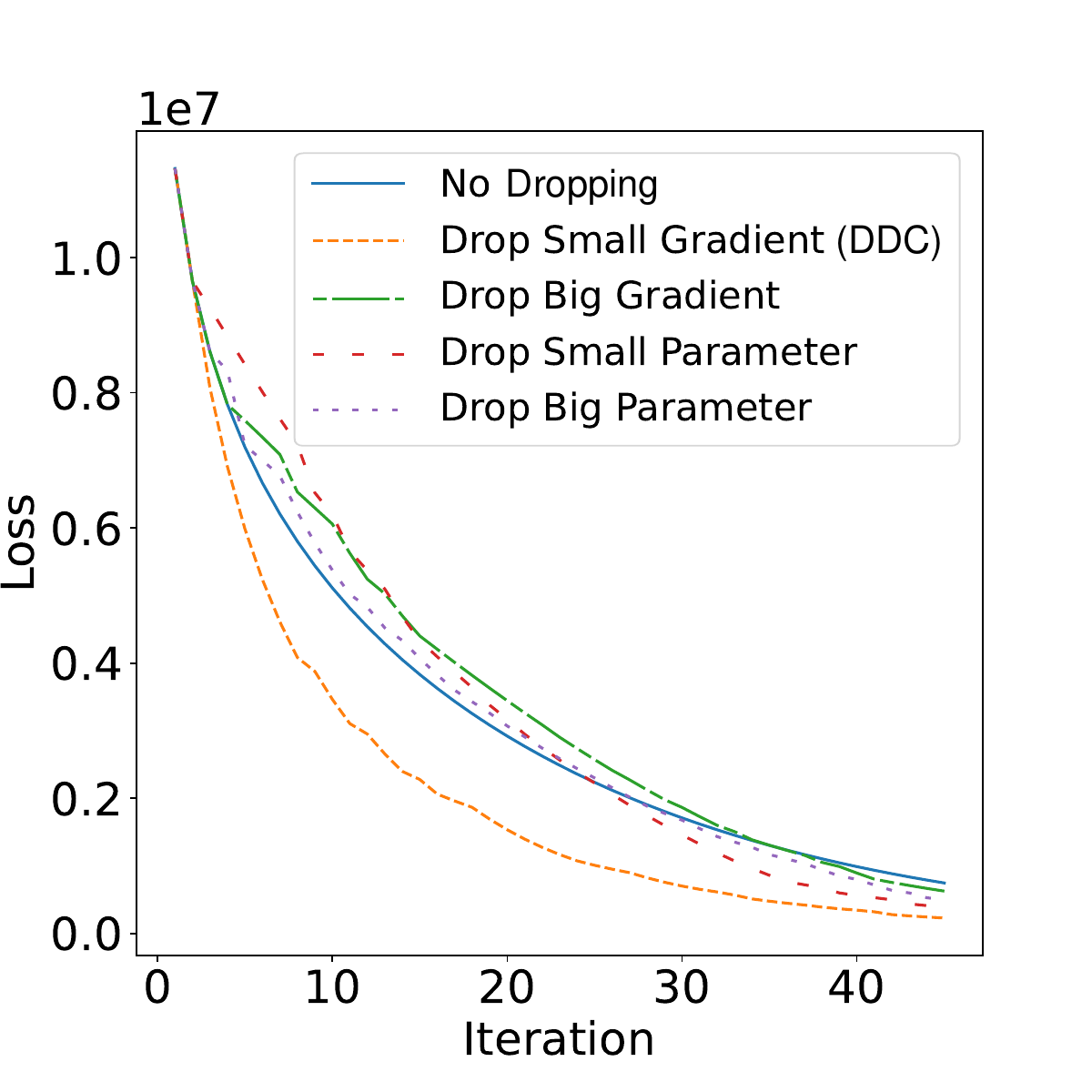}
         \caption{$w_1 = 11.5, w_2 = -10$}
         \label{fig:syn-loss-vs-iter-1}
     \end{subfigure}
     \begin{subfigure}{0.48\textwidth}
         \centering         \includegraphics[width=\textwidth]{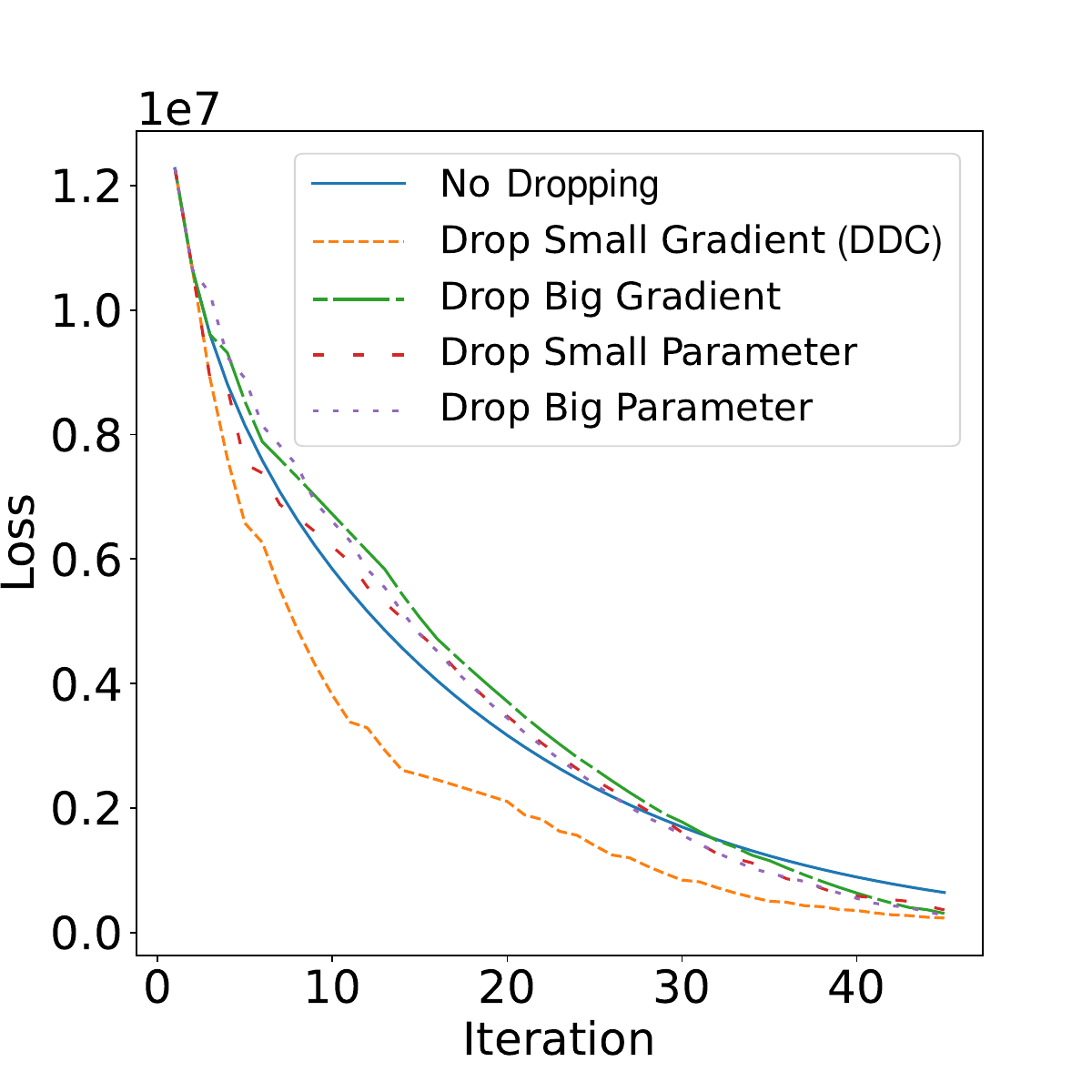}
         \caption{$w_1 = 9, w_2 = -10.5$}
         \label{fig:syn-loss-vs-iter-2}
     \end{subfigure}
        \caption{Loss vs.~epochs of different dropping strategies.}
        \label{fig:syn-loss-vs-iter}
\end{figure}

We generate a synthetic dataset to analyze the learning process of various edge-dropping strategies. For each instance $i$, we sample two independent features $x_1^{(i)}$ and $x_2^{(i)}$, each $x_j^{(i)} \sim N(0,1)$.  We fix the values of $w_1$ and $w_2$ and let the target $y^{(i)}$ be $w_1 x_1^{(i)} + w_2 x_2^{(i)} + \xi$, where $\xi \sim N(0,1)$.  We only use two parameters, $w_1$ and $w_2$,  because it is easier to visualize the parameter updating/learning process in a 2-dimensional contour plot.

Figure~\ref{fig:linreg-contours} shows the parameter learning process of various edge-dropping strategies: the first column (labeled ``No regularization'') represents the scenario of the no-dropping mechanism. The second column (labeled ``Drop Small Gradient'') corresponds to the DDC method. As for the other methods, ``Drop Big Gradient'' tends to drop the edges with more significant gradients; ``Drop Small Parameter'' favors dropping edges with small weights; ``Drop Big Parameter'' tends to drop edges with large weights. The horizontal and vertical axes denote the values of the estimated $w_1$ and $w_2$ at different epochs, and the contour lines represent the loss. We highlight the initial values of $w_1$ and $w_2$ and their values at the 7th, 13th, and 19th epochs by blue pluses, red stars, red triangles, and red diamonds, respectively. As shown, updating the parameters with DDC reaches small losses faster than the compared baselines. We randomly initialize the values of $w_1$ and $w_2$ and repeat the process several times.   Figure~\ref{fig:linreg-contours} shows two cases with different initializations.

We show the relationship between iteration and the losses of the compared methods in Figure~\ref{fig:syn-loss-vs-iter}. The results show that DDC gets the lowest loss when giving a fixed iteration count. The methodologies ``Drop Small Parameter'', ``Drop Big Parameter'', and ``No Dropping'' are in the middle, and ``Drop Big Gradient'' yields the largest loss given a fixed iteration count. We demonstrate the results using two sets of initial values for $w_1$ and $w_2$. Other sets of initial values give similar patterns. It appears that the parameter values per se are unlikely to be influential factors in deciding the dropping rate; the gradients are more influential.

\subsection{Experiments on Open Datasets}
\label{sec:open-data}

\begin{table}[tb]
\begin{center}  
    \caption{The test accuracies (\%) of applying SimpleCNN on MNIST.  We repeat each experiment 5 times and report the mean $\pm$ standard deviation. Let the test accuracies of ``No Dropping'' and a dropping method $m$ be $a \pm b$ and $c \pm d$, respectively, if $a+b < c-d$, we label the method $m$ with symbol $\upuparrows$ (much better). If $a < c$ but $a + b \nless c - d$, we denote $m$ with $\uparrow$ (better).  If $a-b > c+d$, we denote $\downdownarrows$ (much worse). Finally, if $a > c$ but $a - b \ngtr c + d$, we label $\downarrow$ (worse). We also highlight the method with the largest average accuracy with a boldface.} 
\label{tab:mnist-simple-cnn} 
\begin{tabular}{cl}
\toprule
Method & SimpleCNN \\
\midrule
No Dropping & $99.08 \pm 0.04$ \\
DDC               & $\boldsymbol{99.25 \pm 0.01} (\upuparrows)$ \\
Dropout           & $99.03 \pm 0.08 (\downarrow)$ \\
DropConnect       & $99.22 \pm 0.03 (\upuparrows)$ \\
Standout & $99.01 \pm 0.02 (\downdownarrows)$\\
Drop Small Parameter        & $99.24 \pm 0.01 (\upuparrows)$ \\
Drop Big Parameter          & $99.10 \pm 0.01 (\uparrow)$ \\
Drop Big Gradient         & $99.12 \pm 0.01 (\uparrow)$ \\
\bottomrule
\end{tabular}
\end{center}
\end{table}

The previous section shows that DDC helps linear models learn faster. This section explores DDC's capacity for more complicated deep learning models using four open datasets: MNIST~\cite{lecun2010mnist}, CIFAR-10, CIFAR-100~\cite{krizhevsky2009learning}, and NORB~\cite{lecun2004learning}.

In addition to the baselines introduced in Section~\ref{sec:syn-data}, we include Dropout, DropConnect, and Standout for comparison. We adjust the dropping rate $r^{(l)}$ in Algorithm~\ref{alg:one-layer-forward} to be close to the dropping probability of the compared baselines whenever possible.

We test the convolutional neural network used in the DropConnect paper~\cite{wan2013regularization} (called SimpleCNN below). In addition, we add two more complicated networks --AlexNet~\cite{krizhevsky2017imagenet} and VGG~\cite{simonyan2014very} -- for comparison.

MNIST contains grayscale images; the size of each image is $28 \times 28$. The dataset is simple, so we tested only different dropping strategies on SimpleCNN, which includes three convolutional layers and two fully connected layers. The detailed structure is given in~\cite{wan2013regularization}; the hyperparameters and detailed settings are included in the experimental code.

The results are presented in Table~\ref{tab:mnist-simple-cnn}. After repeating each experiment five times, we report the average accuracy $\pm$ standard deviation. As demonstrated, most test accuracies improve when dropping strategies are applied, and the proposed DDC method achieves the best accuracy among all compared methods. Although all methods exhibit accuracies above $99\%$, DDC is significantly better because it consistently shows very small standard deviations across all repeated trials.

\begin{table}[tb]
\begin{center}
\caption{The test accuracies (\%) of applying AlexNet and VGG on the CIFAR-10 dataset, the symbols $(\upuparrows)$, $(\uparrow)$, $(\downarrow)$, and $(\downdownarrows)$ are the same as in Table~\ref{tab:mnist-simple-cnn}.} 
\label{tab:cifar10-cnn-alexnet-vgg} 
\begin{tabular}{@{}ccc@{}}
\toprule
                     & \multicolumn{2}{c}{Network Structure} \\ \midrule
Method      & AlexNet     & VGG     \\ \midrule
No Dropping         & $81.04 \pm 0.04$ & $90.46 \pm 0.12$ \\
DDC      & $\boldsymbol{84.25 \pm 0.09} (\upuparrows)$ & $\boldsymbol{90.94 \pm 0.11} (\upuparrows)$ \\
Dropout & $83.86 \pm 0.09 (\upuparrows)$ & $90.66 \pm 0.07 (\upuparrows)$ \\
DropConnect  & $83.52 \pm 0.17 (\upuparrows)$ & $90.68 \pm 0.08 (\upuparrows)$ \\
Standout & $83.75 \pm 0.04 (\upuparrows)$ & $90.64\pm 0.08 (\uparrow)$ \\
Drop Small Parameter & $83.00 \pm 0.10 (\upuparrows)$ & $89.88 \pm 0.01 (\downdownarrows)$ \\
Drop Big Parameter & $83.81 \pm 0.06 (\uparrow)$ & $90.45 \pm 0.02 (\downarrow)$ \\
Drop Big Gradient & $83.19 \pm 0.16 (\uparrow)$ & $90.88 \pm 0.03 (\upuparrows)$ \\ \bottomrule
\end{tabular}
\end{center}
\end{table}

CIFAR-10 is a more challenging dataset: it contains color images; each image's size is $32 \times 32$. We use AlexNet~\cite{krizhevsky2017imagenet} and VGG~\cite{simonyan2014very} network structures for the experiments because CIFAR-10 is more complex than MNIST.

Table~\ref{tab:cifar10-cnn-alexnet-vgg} shows the results of applying AlexNet and VGG using the CIFAR-10 dataset. For both the AlexNet and VGG networks, the dropping methods mostly increase test accuracies, and our DDC performs the best among all methods in both networks.

We also test VGG and AlexNet on CIFAR-100 by changing the output softmax layer to 100 categories; all other settings are identical to the previous setup.

Table~\ref{tab:cifar100-alexnet-vgg} shows the results of applying the AlexNet and VGG on the CIFAR-100 dataset. The results are similar to those reported earlier: methods with dropping rates usually perform better than without dropping. Our DDC performs best on the VGG network and second on AlexNet.

Finally, we apply the AlexNet and the VGG network to the NORB dataset. The outcomes of these experiments are displayed in Table~\ref{tab:norb-alexnet-vgg}. Our DDC method still consistently outperforms the other techniques on both datasets. It is noteworthy that, in some cases when using the AlexNet, methodologies that involve dropping rates occasionally yield lower performance compared to methods that do not employ any dropping strategy on the NORB dataset. However, the difference is not statistically significant. By examining the mean accuracies plus or minus one standard deviation, we observe that the performance of the poorer performing methods with dropping rates overlaps with the ``No Dropping'' method.

\begin{table}[tb]
\centering
\caption{The test accuracies (\%) of applying AlexNet and VGG on the CIFAR-100 dataset, the symbols $(\upuparrows)$, $(\uparrow)$, $(\downarrow)$, and $(\downdownarrows)$ are the same as in Table~\ref{tab:mnist-simple-cnn}.}
\label{tab:cifar100-alexnet-vgg} 
\begin{tabular}{@{}ccc@{}}
\toprule
                     & \multicolumn{2}{c}{Network Structure} \\ \midrule
Method               & AlexNet              & VGG             \\ \midrule
No Dropping & $62.53 \pm 0.50$ & $71.38 \pm 0.03$ \\
DDC & $64.06 \pm 0.37 (\upuparrows)$ & $\boldsymbol{72.09 \pm 0.04} (\upuparrows)$ \\
Dropout & $ \boldsymbol{66.53 \pm 0.27} (\upuparrows)$ & $71.69 \pm 0.06 (\upuparrows)$ \\
DropConnect & $63.72 \pm 0.11 (\upuparrows)$ & $71.60 \pm 0.21 (\uparrow)$ \\
Standout & $63.97 \pm 0.38 (\upuparrows)$ & $71.53 \pm 0.31 (\uparrow)$ \\
Drop Small Parameter & $63.10 \pm 0.13 (\uparrow)$ & $71.71 \pm 0.10 (\upuparrows)$ \\
Drop Big Parameter & $63.02 \pm 0.47 (\uparrow)$ & $71.82 \pm 0.03 (\upuparrows)$ \\
Drop Big Gradient & $62.92 \pm 0.22 (\uparrow)$ & $71.13 \pm 0.07 (\downdownarrows)$ \\ \bottomrule
\end{tabular}
\end{table}

\begin{table}[tb]
\centering
\caption{The test accuracies (\%) of applying AlexNet and VGG on the NORB dataset, the symbols $(\upuparrows)$, $(\uparrow)$, $(\downarrow)$, and $(\downdownarrows)$ are the same as in Table~\ref{tab:mnist-simple-cnn}.}
\label{tab:norb-alexnet-vgg}
\begin{tabular}{@{}ccc@{}}
\toprule
                     & \multicolumn{2}{c}{Network Structure} \\ \midrule
Method               & AlexNet             & VGG             \\ \midrule
No Dropping & $92.08 \pm 0.11$ & $93.36 \pm 0.03$ \\
DDC & $\boldsymbol{92.93 \pm 0.09} (\upuparrows)$ & $\boldsymbol{94.20 \pm 0.16} (\upuparrows)$ \\
Dropout & $92.04 \pm 0.07 (\downarrow)$ & $93.79 \pm 0.25 (\upuparrows)$ \\
DropConnect & $92.09 \pm 0.14 (\uparrow)$ & $93.94 \pm 0.19 (\upuparrows)$ \\
Standout & $92.08 \pm 0.07$ & $93.81 \pm 0.08 (\upuparrows)$ \\
Drop Small Parameter & $92.42 \pm 0.10 (\upuparrows)$ & $94.02 \pm 0.05 (\upuparrows)$ \\
Drop Big Parameter & $92.02 \pm 0.04 (\downarrow)$ & $93.82 \pm 0.15 (\upuparrows)$ \\
Drop Big Gradient & $91.85 \pm 0.13 (\downarrow)$ & $94.17 \pm 0.07 (\upuparrows)$ \\ \bottomrule
\end{tabular}
\end{table}

\section{Conclusion} \label{sec:conc}

This paper presents a straightforward yet effective methodology for dynamically adjusting the dropping rate of the edges in a neural network. This approach avoids the complexity of introducing additional learning parameters and simplifies the implementation process. We demonstrate the parameter updating process through a series of experiments using a synthetic dataset and multiple open datasets and compare our proposed methodology against Dropout and its various adaptations. The results of these experiments indicate that our method surpasses traditional approaches in nearly all scenarios.

The efficacy of gradient magnitude as a reliable indicator for setting the dropping rate has been validated through our results, suggesting a promising direction for future enhancements. Building on this foundation, we propose to develop a model that algorithmically uses gradients as features to determine optimal dropping rates. We hypothesize that this advanced strategy could potentially elevate the model's predictive accuracy even further. However, it's important to note that this approach would introduce additional parameters that require learning, which could extend the training duration and increase computational demands.

Furthermore, we are interested in delving into a more rigorous theoretical analysis of our methodology. Previous studies, such as those referenced in \cite{gal2016dropout}, have explored the connection between Dropout techniques and Bayesian learning. Inspired by this, we are interested in investigating potential theoretical links between our dynamic edge dropping approach and Bayesian inference principles. This exploration could offer deeper insights into the probabilistic foundations of our method and possibly reveal new theoretical reasoning that could explain why our Dynamic DropConnect method enhances model robustness and generalizability more effectively than traditional Dropout techniques.

In summary, our current methodology provides a significant improvement over traditional techniques. Our future efforts will focus on refining this approach by incorporating more sophisticated learning strategies and gaining deeper theoretical insight. Our goal is to further boost the robustness and accuracy of neural network training.

\section*{Acknowledgement}
We acknowledge partial support from the National Science and Technology Council of Taiwan under grant number 113-2221-E-008-100-MY3. We thank to National Center for High-performance Computing (NCHC) of National Applied Research Laboratories (NARLabs) in Taiwan for providing computational and storage resources.

\bibliographystyle{unsrt}  
\bibliography{ref}

\end{document}